\documentclass{article}
\usepackage{ijcai25}

\usepackage{times}
\usepackage{soul}
\usepackage{url}
\usepackage[hidelinks]{hyperref}
\usepackage[utf8]{inputenc}
\usepackage[small]{caption}
\usepackage{graphicx}
\usepackage{amsmath}
\usepackage{amsthm}
\usepackage{booktabs}
\usepackage{algorithm}
\usepackage{algorithmic}
\usepackage[switch]{lineno}
\usepackage{amssymb}
\usepackage{xcolor}
\usepackage{CJKutf8}
\usepackage{pifont}
\usepackage{bbding}
\usepackage{subfigure}
\usepackage{graphicx}
\usepackage{multirow}

\title{Reliable Disentanglement Multi-view Learning Against View Adversarial Attacks}

\author{
Xuyang Wang$^1$
\and
Siyuan Duan$^1$\and
Qizhi Li$^1$\and
Guiduo Duan$^2$\and
Yuan Sun$^{1,3}$\footnotemark[1]\And
Dezhong Peng$^{1,4,5}$\footnotemark[1]\\
\affiliations
$^1$College of Computer Science, Sichuan University, China\\
$^2$Laboratory of Intelligent Collaborative Computing, University of Electronic Science and Technology of China, China\\
$^3$National Key Laboratory of Fundamental Algorithms and Models for Engineering Numerical Simulation, Sichuan University, China\\
$^4$Tianfu Jincheng Laboratory, China\\
$^5$Sichuan National Innovation New Vision UHD Video Technology Co., Ltd., China\\
\emails
xywang@stu.scu.edu.cn,
\{ddzz122773315, mrqz945, sunyuan\_work\}@163.com,
guiduo.duan@uestc.edu.cn,
pengdz@scu.edu.cn
}

\begin{document}

\maketitle

\footnotetext[1]{Corresponding authors}

\begin{abstract}
Trustworthy multi-view learning has attracted extensive attention because evidence learning can provide reliable uncertainty estimation to enhance the credibility of multi-view predictions. Existing trusted multi-view learning methods implicitly assume that multi-view data is secure. However, in safety-sensitive applications such as autonomous driving and security monitoring, multi-view data often faces threats from adversarial perturbations, thereby deceiving or disrupting multi-view models. This inevitably leads to the adversarial unreliability problem (AUP) in trusted multi-view learning. To overcome this tricky problem, we propose a novel multi-view learning framework, namely Reliable Disentanglement Multi-view Learning (RDML). Specifically, we first propose evidential disentanglement learning to decompose each view into clean and adversarial parts under the guidance of corresponding evidences, which is extracted by a pretrained evidence extractor. Then, we employ the feature recalibration module to mitigate the negative impact of adversarial perturbations and extract potential informative features from them. Finally, to further ignore the irreparable adversarial interferences, a view-level evidential attention mechanism is designed. Extensive experiments on multi-view classification tasks with adversarial attacks show that RDML outperforms the state-of-the-art methods by a relatively large margin. Our code is available at \url{https://github.com/Willy1005/2025-IJCAI-RDML}.
\end{abstract}

\section{Introduction}
In practical scenarios, an object can often be described by multiple views of different feature types and modalities, which leads to a growing interest in multi-view learning \cite{mv9,kg,mv11,mv12}. Thanks to the power of deep learning, deep multi-view learning has exhibited remarkable advantages by integrating and mining both valuable complementary and consistency information of multi-views \cite{mv2,mv3,mv7}. Thus, in recent years, multi-view learning has attracted widespread attention \cite{qmf,mv13,mv14}. For example, predictive dynamic fusion (PDF) \cite{pdf} proposes an intuitive and rigorous multimodal fusion paradigm from the perspective of generalization error.

Although these above methods have achieved pleasing performance, their results could be uncertain and unreliable due to the attribute differences and heterogeneity of multi-view data. This greatly limits the application of multi-view learning in various fields, especially medical diagnosis or autonomous driving. To this end, a new trusted learning paradigm for multi-view classification is proposed to enhance trusted decisions. For instance, Trusted Multi-view Classification (TMC) \cite{enn7} introduces the evidence theory to construct the Dirichlet distribution, thereby providing uncertainty estimation for multi-view decisions to enhance reliability. To make reliable decisions under noise labels, TMNR \cite{enn3} proposes trusted multi-view noise refining to overcome the negative effects of noisy labels. 

\begin{figure}[t]
\centering
\subfigure[Uncertainty density]{\includegraphics[width=0.46\linewidth]{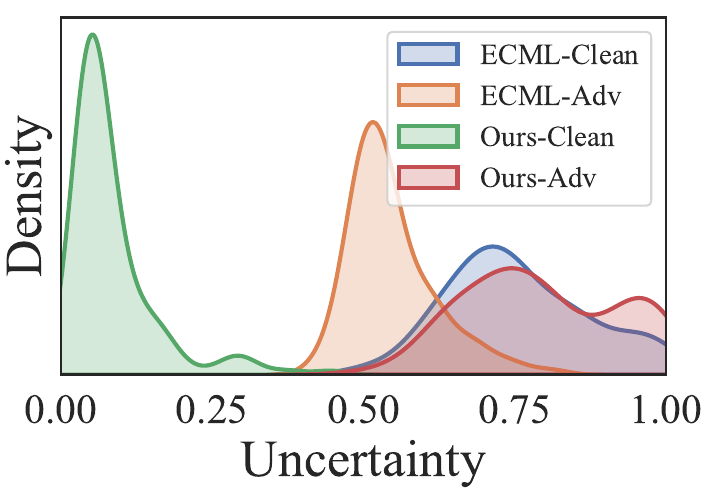}}
\subfigure[Classification accuracy]{\includegraphics[width=0.52\linewidth]{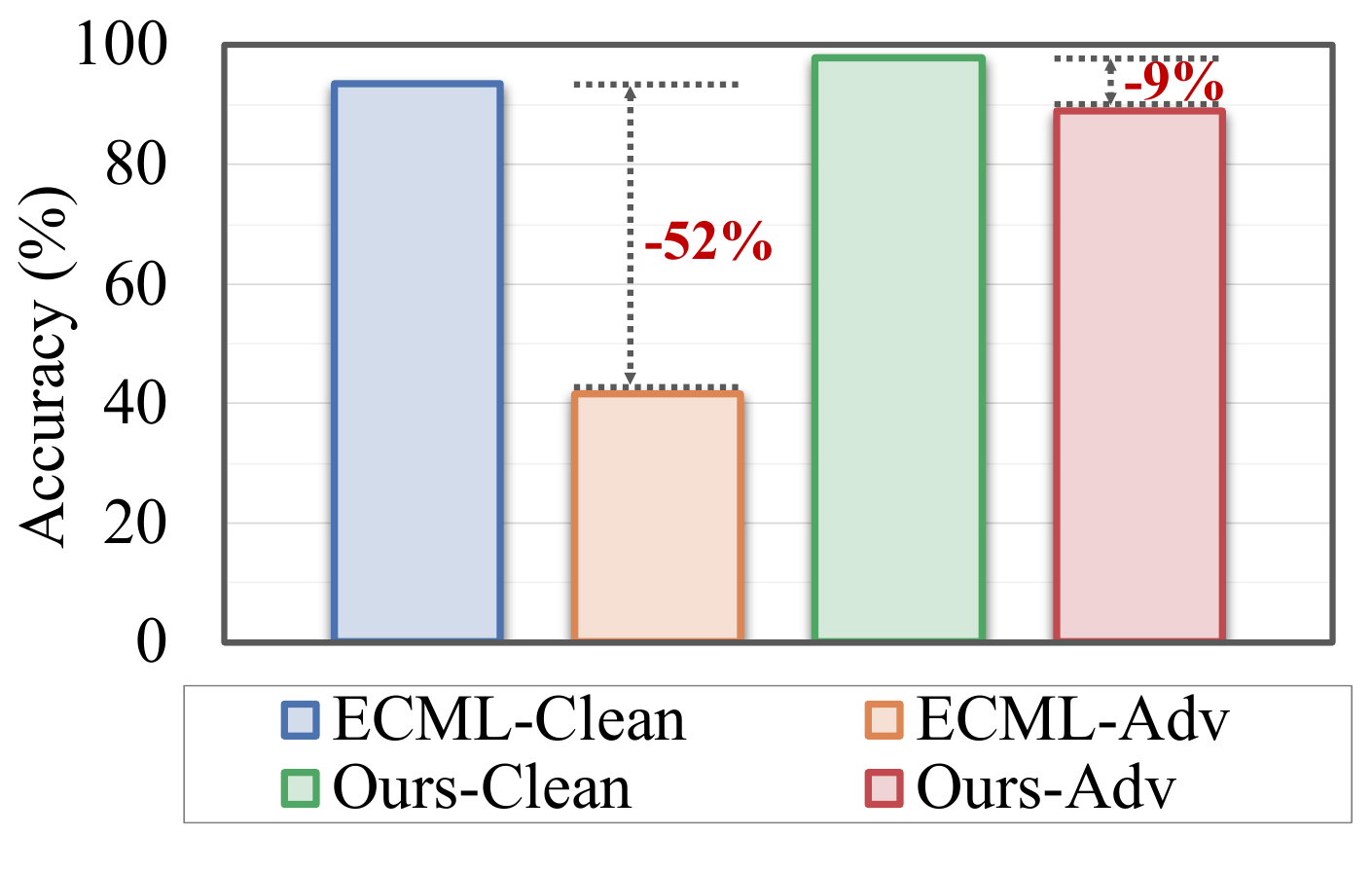}}
\caption{We conduct experiments on the PIE dataset in clean and adversarial settings, and show a toy example of the AUP. Note that only one view is randomly attacked via PGD. (a) presents the estimated uncertainties of the ECML method and our RDML. (b) shows the classification accuracy of these methods.
}
\label{fig:intro}
\end{figure}

Almost all existing trusted multi-view learning methods implicitly assume that multi-view data is secure \cite{mv10}. In practice, however, in safety-critical applications such as autonomous driving and security monitoring, multi-view data could be susceptible to adversarial attacks, which can deceive or disrupt multi-view models. This vulnerability inevitably leads to the adversarial unreliability problem (AUP) in trusted multi-view learning. As shown in Fig.\ref{fig:intro}, adversarial attacks (such as projected gradient descent attack \cite{atk1}) are imposed on the multi-view data. From the figure, we can observe that, after being subjected to attacks, even if only one view is attacked, the state-of-the-art evidence-based method (i.e., ECML) still shows a significant decline in classification accuracy. Worse still, instead of increasing with the substantial performance decline, the estimated uncertainties are significantly lower than those in the clean setting. This indicates that the evidence-based uncertainty estimation mechanism fails under adversarial perturbations, thereby leading to the AUP.

\begin{figure*}
\centering
\includegraphics[width=0.9\textwidth]{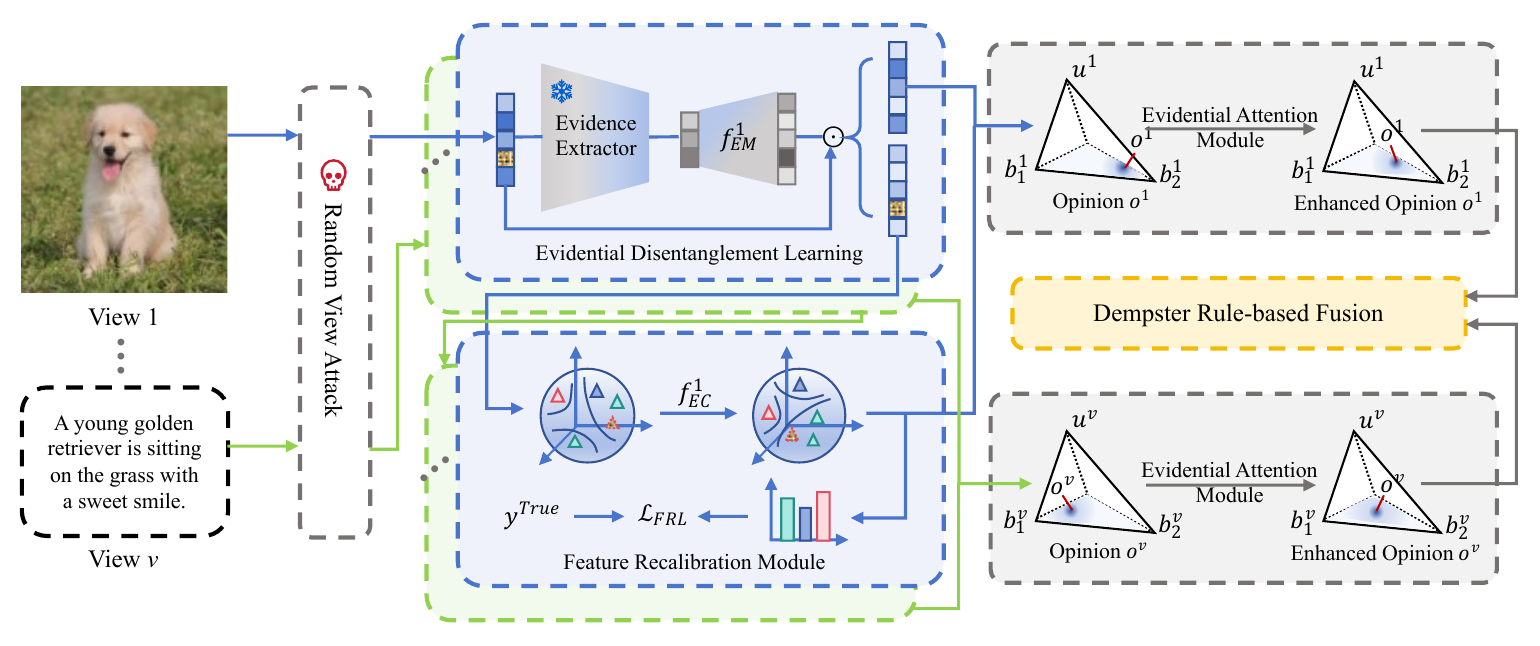}
\caption{The framework of RDML. (i) Evidential disentanglement learning uses the pretrained evidence extractor $E_{pt}(\cdot)$ to conduct a robustness analysis of features under random view attacks, and generate a robustness mask to decouple clean and adversarial features. (ii) The feature recalibration module will rectify the adversarial features. For features that are difficult to repair, RDML will generate evidential attention with the guidance of $E_{pt}(\cdot)$ to further mitigate the interference of adversarial features. (iii) RDML introduces the Dempster Rule-based Fusion for opinion aggregation.}
\label{model}
\end{figure*}

To overcome the above AUP, we propose a novel multi-view learning framework, namely Reliable Disentanglement Multi-view Learning (RDML). In the first stage, we present a perturbation-insensitive pretraining scheme to train an evidence extractor, thereby providing reliable category-level evidence and uncertainty estimation in subsequent stages. In the second stage, as shown in Fig.\ref{model}, our method mainly consists of three key modules: the evidential disentanglement learning module, the feature recalibration module, and the evidential attention module. To be specific, RDML first proposes evidential disentanglement learning to decompose each view into clean and adversarial features under the guidance of pretrained corresponding evidence. Then, to prevent the negative effect of the adversarial features, we propose feature recalibration to recalibrate these feature units for additional discriminative information, thereby obtaining more robust features. Finally, to reduce the interference of stubborn adversarial features that are difficult to calibrate, we design a view-level evidential attention mechanism to enhance the robustness against adversarial features. The main contributions of this paper are as follows.
\begin{itemize}
\item This paper studies a less-touched adversarial unreliability problem (AUP) in trusted multi-view learning and proposes a new Reliable Disentanglement Multi-view Learning (RDML) framework against view adversarial attacks. To the best of our knowledge,  for the first time, we address the AUP caused by view adversarial attacks.
\item We propose evidential disentanglement learning to guide the stripping of adversarial information from multi-view representations using a pretrained evidence model. To mitigate the interference of perturbations, we propose feature recalibration to rectify the weak adversarial features, and further present evidence attention to deal with the stubborn adversarial features.
\item We conduct extensive experiments on six multi-view datasets to verify the effectiveness and robustness of our RDML under both adversarial and clean conditions.
\end{itemize}

\section{Related Work}
\subsection{Trusted Multi-view Learning}
Deep multi-view learning utilizes view-specific deep representations to enhance the integration and understanding of multi-source information \cite{mv4,mv6,mv5,mv8}. Han et al. point out that existing studies have placed too much emphasis on improving the performance of deep multi-view learning methods in various scenarios while neglecting to enhance the reliability of multi-view decision \cite{enn1}. They then propose the Trusted Multi-view Classification method. TMC transforms traditional classification networks into evidential neural networks (by replacing the last softmax function of the classification networks with Relu function to ensure that each output value is non-negative) and uses the extracted evidence representations to model the Dirichlet distribution \cite{edl1,edl2}. The Dirichlet distribution can output classification category probability and uncertainty, thereby achieving reliable multi-view decision. Finally, TMC introduces Dempster rule to fuse multi-view opinions. 

Xu et al. introduce evidence learning into conflictive multi-view learning, extracting view-specific opinion through the parameterized Dirichlet distribution \cite{enn2}. And a conflictive opinion aggregation method is designed for multi-view fusion. Liu et al. propose an opinion fusion method based on evidence accumulation, in which the evidence representations of different views are accumulated to obtain the overall opinion \cite{enn6}. Yue et al. discover the vulnerability of trusted multi-view approaches to adversarial examples and attributed this vulnerability to the difficulty of accurately assessing the quality of adversarial examples \cite{enn10}. In this paper, we address the AUP from a more fundamental perspective. That is, we first enhance the adversarial insensitivity of the evidence neural network (perturbation-insensitive pretraining), and subsequently minimize the interference and harm that adversarial perturbations inflict on multi-view fusion and decision to the greatest extent possible (evidence-based disentanglement, feature recalibration and view-level evidential attention).
\subsection{Deep Adversarial Defense}
Improving the robustness of deep neural networks against adversarial perturbations has long been a goal in deep learning community \cite{atk2,atk3,atk7}. Since there are different defense strategies for different types of attacks, here we mainly focus on deep defense methods against white-box attacks. In recent years, adversarial training has been widely proven to be effective in enhancing the adversarial robustness of neural networks \cite{atk4,atk5}. By using both clean samples and adversarial samples as training data, the insensitivity of neural networks to adversarial perturbations can be enhanced.

Moreover, since disentanglement learning is good at separating information, it is naturally suitable for dealing with perturbed samples \cite{atk6,atk8,atk9}. Typically, disentanglement learning decomposes perturbed samples into clean and adversarial features to mitigate the interference caused by adversarial perturbations. However, existing disentanglement learning methods lack the support of effective clues and trusted guidance, and thus the adversarial robustness is limited. In this paper, we propose an evidence-based disentanglement method to resist adversarial perturbations. The disentanglement is guided by a pretrained evidence extractor, so it can improve the reliability and adversarial robustness of the evidential networks.

\section{Method}
\subsection{Problem Definition}
Suppose that there is a multi-view dataset $D_{N}^{V}$ with $N$ instances and $V$ views, and $x_{n}^{v}\in\mathbb{R}^{d_{v}}$ ($n=1,2,...,N$) is an instance from $D_{N}^{V}$. ${d_{v}}$ is the dimension of the $v$-th view. The corresponding class labels are $\left\{\mathbf{y}_n\right\}_{n=1}^N$. The number of classes is $K$. The random view adversarial attacks are conducted to each multi-view instance by the following formula,
\begin{equation}
\widehat{x}_{n}^{v}=\text{AdvAttack}(x_{n}^{v}),\label{eq:atk}
\end{equation}
where ${d_{v}}$ is the dimension of the $v$-th view. $\widehat{x}_{n}^{v}$ means the adversarial version of ${x}_{n}^{v}$. Our goal is to learn a robust evidential model against view adversarial attacks.

\subsection{Overview}
When facing adversarial attacks, existing trusted multi-view methods can easily lead to the AUP, thereby weakening the performance of multi-view learning models. To address the above problem, in this paper, we propose Reliable Disentanglement Multi-view Learning (RDML) against view adversarial attacks, which consist of two stages. In the first stage, we propose a perturbation-insensitive pretraining scheme to enhance the stability and adversarial insensitivity of evidential neural networks. To be specific, this approach introduces random view attacks into the pretraining of multi-view evidential networks, thereby making the evidence extractor provide the support of reliable category-level evidence and uncertainty estimation. The objective function could be formulated as follows:
\begin{equation}
    \mathcal{L}_{PT}=\mathcal{L}_{ECL}(\alpha_{n})+\sum_{v=1}^{V}\mathcal{L}_{ECL}(\alpha_{n}^{v})+\mathcal{L}_{ACL},
\end{equation}
where the $\mathcal{L}_{ECL}(\alpha_{n})$ is the evidential classification loss; $\alpha_{n}$ is the Dirichlet parameter; $\mathcal{L}_{ACL}$ denotes the adversarial consistency loss.

In the second stage, RDML first proposes evidential disentanglement learning. To be specific, we use the pretrained evidence extractor to analyze the features to be decoupled and map the extracted category-level evidence into a robustness-aware soft mask. The higher the score of the mask for a feature, the more likely the feature is to be a robust (or clean) feature, and vice versa, it is more likely to be an adversarial feature. Afterward, we utilize the mask to separate clean and adversarial features.  In addition, we believe that some weak adversarial features could easily be converted into clean features.  To this end, we propose feature recalibration to correct these adversarial features into clean features. For the remaining part of stubborn adversarial features, we design a view-level evidential attention mechanism to reduce the interference of these adversarial features that are difficult to correct, thereby enhancing the robustness against adversarial features. The objective function could be expressed as 
\begin{equation}
\begin{aligned}
    \mathcal{L_{T}}=&\mathcal{L}_{ECL}(\alpha_{n})+\sum_{v=1}^{V}\mathcal{L}_{ECL}(\alpha_{n}^{v})+\mathcal{L}_{ACL}\\&+\gamma_{1}\mathcal{L}_{EDL}+\gamma_{2}\mathcal{L}_{FRL},
\end{aligned}
\end{equation}
where $\mathcal{L}_{EDL}$ and $\mathcal{L}_{FRL}$ represent evidential disentanglement loss and feature recalibration loss, respectively; $\gamma_{1}$ and $\gamma_{2}$ are two balancing parameters.

\subsection{Perturbation-insensitive Pretraining}
Although evidential neural networks can provide corresponding uncertainties to enhance the ability of trusted decisions, they could become unstable and even ineffective after being subjected to adversarial attacks. To overcome this issue, we propose a perturbation-insensitive pretraining strategy that incorporates adversarial samples to enhance the adversarial insensitivity of the evidence extractor $E_{pt}(\cdot)=\{E_{pt}^{v}(\cdot)\}_{v=1}^{V}$, thereby providing robust evidence support. 
To be specific, for adversarial multi-view data, to enhance the robustness of the evidence extractor for adversarial attacks, we then mix the adversarial multi-view samples with clean samples and utilize the evidence extractor for the mixed samples $x_{mix}$ to extract the corresponding classification evidence $e$, i.e.,
\begin{equation}
    e_{k}^{v}=E_{pt}^{v}(x_{mix}^{v}), k=1,2,...,K,
\end{equation}
where $K$ denotes the number of classes. Afterward, we obtain the parameter $\alpha_{k}^{v}=e_{k}^{v}+1$ required for modeling the Dirichlet distribution based on the evidence. The view-specific opinion $o^{v}=(b^{v},u^{v})$ are also obtained according to following formula,
\begin{equation}
    b_{k}^{v}=\frac{e_{k}^{v}}{S^{v}},u^{v}=\frac{K}{S^{v}},\label{eq:bu}
\end{equation}
where $b\geq0$ and $u\geq0$ denote the belief mass and uncertainty ($\sum_{k=1}^{K}b_{k}^{v}+u^{v}=1$), and $S^{v}=\sum_{k=1}^{K}\alpha^{v}_{k}$ represents the Dirichlet strength. Given the $S^{v}$ and $\alpha_{k}^{v}$, the probability of the $v$-th view for the $k$-th class is $p_{k}^{v}=\frac{\alpha_{k}^{v}}{S^{v}}$. After that, Dempster rule is used to combine multi-view opinions, and the joint opinion is $o=o^{1}\otimes o^{2} \otimes ...\otimes o^{v}$, where $\otimes$ is the Dempster rule based fusion operation.

To optimize the evidence extractor, following \cite{enn1}, we use the evidential classification loss $\mathcal{L}_{ECL}(\alpha_{n})$, i.e.,
\begin{equation}
    \mathcal{L}_{ECL}(\alpha_{n})=\mathcal{L}_{ECE}+KL[D(p_{n}\mid\tilde{\alpha}_{n})\parallel D(p_{n}\mid1)],
\end{equation}
\begin{equation}
    \mathcal{L}_{ECE}(\alpha_{n})=\sum_{k=1}^{K}y_{nk}(\psi(S_{n})-\psi(\alpha_{n})),
\end{equation}
where $\psi(\cdot)$ denotes the digamma function; $\tilde{\alpha}_{n}=y_{n}+(1-y_{n})\odot\alpha_{n}$ is the adapted Dirichlet parameter avoiding penalizing the evidence of the correct category to zero. $\mathcal{L}_{ECE}(\alpha_{n})$ represents the evidential cross-entropy loss, which requires the model to extract more evidence for the correct category compared to other categories; and the Kullback-Leibler (KL) divergence restricts the model to extract as little evidence as possible from the incorrect categories. 

Besides, since the instances in AMVL are subjected to random view attacks, we hope to enhance model robustness against adversarial attacks by constraining the differences among the predicted probability distributions of different views. Therefore, we construct the following adversarial consistency loss $\mathcal{L}_{ACL}$, i.e.,
\begin{equation}
    \mathcal{L}_{ACL}=\frac{1}{V-1}\sum_{v_{1}=1}^{V}(\sum_{v_{2}\neq v_{1}}^{V}\frac{\sum_{k=1}^{K}|p_{k}^{v_{1}}-p_{k}^{v_{2}}|}{2}),
\end{equation}
Therefore, the well-trained evidence extractor will guarantee robust and stable evidence in subsequent steps with frozen parameters.

\subsection{Evidential Disentanglement Learning}
Although existing disentanglement learning methods can be used for decomposing adversarial and clean features by various meticulously designed losses, the lack of support for effective clues and trusted guidance hinders their performance. To improve the credibility of separating adversarial features, we introduce the evidential learning theory to naturally provide evidence and uncertainty for the decision process. This characteristic can significantly promote reliable feature decomposition. To this end, we design an evidence disentanglement learning module.

In the training stage, given a clean multi-view instance $x=\{x^{v}\in\mathbb{R}^{d_{v}}\}_{v=1}^{V}$, the random view attack is conducted on it, and its adversarial version is $\widehat{x}$. Different from existing disentanglement learning methods, we leverage the pre-trained evidence extractor $E_{\text{pt}}(\cdot)$ to extract adversarially-insensitive evidence from the adversarial samples. Due to strong adversarial insensitivity, pretrained evidence extractor $E_{pt}(\cdot)$ can extract effective evidence under adversarial perturbation, i.e.,
\begin{equation}
    em^{v}=E_{pt}^{v}(\widehat{x}^{v}),
\end{equation}
where $em^{v}\in \mathbb{R}^{K}$ is the evidential map, which implies the category-aware evidence of the $v$-th view of adversarial sample $\widehat{x}^{v}$. Since $em^{v}$ is category-level evidence, we construct a multi-layer perceptron (MLP) based evidence mapping layer $f_{EM,1}(\cdot)=\{f_{EM,1}^{v}(\cdot)\}_{v=1}^{V}$ to map the category-level evidence $em^{v}$ into a feature-level robustness map $rm^{v}=f_{EM,1}^{v}(em^{v})$ ($rm^{v}\in \mathbb{R}^{d_{v}}$), which indicates the amount of evidence for features in each dimension. Features containing more evidence are regarded as clean features, while those containing less are considered adversarial features. Subsequently, in order to facilitate the decomposition of these two types of features, we introduce Gumbel softmax \cite{gbsm} to convert the robustness map into a soft mask score $m^{v}$, i.e.,
\begin{equation}
m^{v}=\frac{e^{(\log{(\sigma(rm^{v}))}+q_1)/\mu}}{e^{(\log{(\sigma(rm^{v}))}+q_1)/\mu}+e^{(\log{(1-\sigma(rm^{v}))}+q_2)/\mu}},
\end{equation}
where each value is a non-negative value less than 1. $\sigma(\cdot)$ represents a Sigmoid function for normalization; $q_{1}$ and $q_{2}$ are two instances sampled from Gumbel distribution (given $u\sim$ Uniform(0, 1), $q=-\log(-\log(u))$); $\mu$ is a temperature coefficient. $m^{v}\in\mathbb{R}^{d_{v}}$ is a feature-level mask score and each dimension of the feature has a score ranging from 0 to 1, a higher score indicates a higher probability of being a clean feature. 

Therefore, we can decompose clean and adversarial features via the following simple feature-level multiplication,
\begin{equation}
    h^{v}_{c}=x^{v}\odot m^{v},
\end{equation}
\begin{equation}
    h^{v}_{a}=x^{v}\odot (1-m^{v}),
\end{equation}
where $h^{v}_{c}$ and $h^{v}_{a}$ are the clean feature of $x^{v}$ and the adversarial one, respectively.

To optimize our evidential disentanglement learning, an evidence disentanglement loss is designed. Evidential disentanglement learning utilizes the evidence output by a pre-trained evidence extractor to generate a soft robustness mask. Therefore, we expect that the distribution of the decoupled clean features is as close as possible to the real distribution, while the distribution of adversarial features is in contrast. Given clean and adversarial feature of the $n$-th instance $h_{c,n}$, $h_{a,n}$, the evidence disentanglement loss can be written as
\begin{equation}
    \mathcal{L}_{EDL}=-\sum_{v=1}^{V}(y_{n}\odot \log(p_{c,n}^{v})+\widehat{y}_{n}\odot \log(p_{a,n}^{v})),
\end{equation}
where $y_{n}$ is the ground truth label of $h_{n}$; $\widehat{y}_{n}$ is the label of a wrong class for $h_{n}$; $p_{c,n}^{v}=f_{EC}^{v}(h_{c,n}^{v})$, $p_{a,n}^{v}=f_{EC}^{v}(h_{a,n}^{v})$ are the classification probabilities of $h_{c,n}^{v}$ and $h_{a,n}^{v}$; $f_{EC}(\cdot)=\{f_{EC}^{v}(\cdot)\}_{v=1}^{V}$ is a group of evidential classifiers where an activation function like Relu is added after each classifier. Note that $f_{EC}(\cdot)$ only participates in the training phase and is not utilized in the pretraining and testing phase.

\subsection{Feature Recalibration}
For the adversarial features, we believe that a part of them can be transformed from the clean features with relatively poor robustness, and this part of features also easily has the potential to be recovered back into clean features. Therefore, we construct an MLP based feature recalibration layer $f_{FC}(\cdot)=\{f_{FC}^{v}(\cdot)\}_{v=1}^{V}$ to rectify weak adversarial features to clean and informative features,  
\begin{equation}
    h^{v}_{cr}=f_{FC}^{v}(h^{v}_{a}),
\end{equation}
where $h^{v}_{cr}$ represents the corrected feature of $\widehat{x}^{v}$. Then we can get the final feature $h^{v}_{f}=h^{v}_{c}+h^{v}_{cr}$. For the feature recalibration module, we expect that the corrected representations can provide as much informative and valuable features as possible for classification. Therefore, the predicted probability distribution of the corrected representations is required to be as close as possible to the ground truth distribution. Thus, we can have the following loss, i.e.,
\begin{equation}
    \mathcal{L}_{FRL}=-\sum_{v=1}^{V}y_{n}\odot \log(f_{EC}^{v}(h_{cr,n}^{v})).
\end{equation}

\subsection{Evidential Attention}
Though decoupling adversarial features and repairing weak adversarial features can alleviate the interference of adversarial perturbations on evidential neural networks to some extent, the impairment brought about by stubborn adversarial features remains difficult to mitigate effectively. Thus, we propose a view-level evidential attention mechanism that generates the evidential attention score by conducting a robust analysis of view features using the pre-trained $E_{pt}(\cdot)$. This mechanism guides the model to focus on the informative clean features and ignore the interference of strong adversarial features that are difficult to utilize. Specifically, we utilize the pretrained evidence extractor $E_{pt}(\cdot)$ to conduct evidence analysis on the feature $h_{f}$. The extracted evidence is transformed into evidential attention scores via a softmax function. Benefiting from the knowledge of the pre-trained evidence extractor, these attention scores imply the robustness of each dimension of $h_{f}$, distinguishing clean, weak adversarial, and hard adversarial features in the form of scores. Since the evidence extractor outputs category-level scores, we map them into feature-level attentions through the evidence mapping layer $f_{EM,2}(\cdot)=\{f_{EM,2}^{v}(\cdot)\}_{v=1}^{V}$, i.e.,
\begin{equation}
    att^{v}_{i}=\frac{e^{E_{pt}^{v}(h_{f,i}^{v})}}{\sum_{j=1}^{K}e^{E_{pt}^{v}(h_{f,j}^{v})}},
\end{equation}
\begin{equation}
    att^{v}=f_{EM,2}^{v}(att^{v}),
\end{equation}
$h_{f,i}^{v}$ denotes the $i$-th element of $h_{f}^{v}$; $att^{v}\in \mathbb{R}^{d_{v}}$ is the view-level evidential attention for $h^{v}_{f}$. The augmented feature is obtained via $h_{aug}^{v}=h_{f}^{v}\odot att^{v}$.

\subsection{Trusted Multi-view Fusion}
Followed by \cite{enn1}, Dempster combination rule is introduced for multi-view fusion. Given two augmented features of two views $h_{aug}^{1}$, $h_{aug}^{2}$, the corresponding evidences $e^{1}=E_{c}^{1}(h_{aug}^{1})$, $e^{2}=E_{c}^{2}(h_{aug}^{2})$ can be extracted via a evidence extractor $E_{c}(\cdot)=\{E_{c}^{v}(\cdot)\}_{v=1}^{V}$. It is worth mentioning that the parameters of $E_{c}(\cdot)$ are copied from the pretrained $E_{pt}(\cdot)$, so as to improve the overall training efficiency. Then two opinions $o^{1}=(b^{1},u^{1})$ and $o^{2}=(b^{2},u^{2})$ are constructed via Eq. (\ref{eq:bu}). After that, we have joint opinion $o=o^{1}\otimes o^{2}=(b,u)$ where
\begin{equation}
b_{k}=\frac{b_{k}^{1}b_{k}^{2}+b_{k}^{2}u^{1}+b_{k}^{1}u^{2}}{1-M}, \\
u=\frac{u^{1}u^{2}}{1-M}.
\end{equation}
$M=\sum_{i\neq j}b_{i}^{1}b_{j}^{2}$ represents the difference between two opinions. And $\frac{1}{1-M}$ is used for normalization. According to the above combination pattern, we have the joint multi-view opinion $o=o^{1}\otimes o^{2} \otimes ...\otimes o^{v}=(b,u)$. Then the joint evidence $e_{k}=b_{k}\times S$, Dirichlet parameter $\alpha_{k}=e_{k}+1$ and Dirichlet strength $S=\frac{K}{u}$ are obtained based on Eq. (\ref{eq:bu}).

\begin{table*}[t]
    \centering
    \begin{tabular}{ccccccccc}
    \hline
        Methods & Ref. & PIE & Scene & Leaves & NUS-WIDE & MSRC & Fashion \\ 
        \hline
        TMC & ICLR'21 & 91.85$\pm$0.23 & 67.71$\pm$0.30 & 86.81$\pm$2.20 & 35.67$\pm$1.37 & 92.38$\pm$2.78 & 95.40$\pm$0.40 \\ 
        ETMC & TPAMI'22 & 93.75$\pm$1.08 & 71.61$\pm$0.28 & \textbf{98.44$\pm$0.40} & 35.58$\pm$1.10 & 90.48$\pm$3.37 & 96.21$\pm$0.36 \\ 
        DUANets & AAAI'21 & 90.59$\pm$1.99 & 51.08$\pm$1.27 & 84.69$\pm$1.06 & 29.38$\pm$1.09 & 77.62$\pm$4.15 & 90.49$\pm$0.97 \\ 
        MMD & CVPR'22 & \underline{94.41$\pm$2.01} & 65.72$\pm$1.38 & 69.05$\pm$1.18 & 28.21$\pm$5.11 & \underline{98.10$\pm$2.33} & 9.58$\pm$0.61 \\ 
        QMF & ICML'23 & 88.82$\pm$1.82 & 65.24$\pm$1.80 & 95.19$\pm$1.53 & 42.33$\pm$2.56 & 94.76$\pm$3.81 & 98.81$\pm$0.16 \\ 
        ECML & AAAI'24 & 93.68$\pm$1.51 & \underline{73.20$\pm$2.16} & 94.63$\pm$1.24 & 41.21$\pm$2.10 & 92.38$\pm$2.78 & 95.24$\pm$0.17 \\ 
        TMNR & IJCAI'24 & 89.71$\pm$1.61 & 66.24$\pm$2.05 & 89.31$\pm$1.80 & 36.75$\pm$1.71 & 90.00$\pm$1.78 & 94.31$\pm$0.45 \\ 
        PDF & ICML'24 & 90.88$\pm$1.36 & 69.61$\pm$1.72 & 98.00$\pm$0.51 & \underline{43.83$\pm$1.73} & 93.33$\pm$3.50 & \underline{98.85$\pm$0.14} \\ 
        RDML & Ours & \textbf{97.79$\pm$0.81} & \textbf{74.40$\pm$1.90} & \underline{97.94$\pm$1.09} & \textbf{46.67$\pm$1.90} & \textbf{99.52$\pm$0.95} & \textbf{98.96$\pm$0.18} \\
        \hline
        $\bigtriangleup$\% &  & \textcolor{blue}{+3.38} & \textcolor{blue}{+1.20} & \textcolor{purple}{-0.50} & \textcolor{blue}{+2.84} & \textcolor{blue}{+1.42} & \textcolor{blue}{+0.11} \\ 
    \hline
    \end{tabular}
    \caption{Classification accuracy (\%) on clean data, where the best and second best results are bolded and underlined, respectively. $\bigtriangleup$\% denotes the improvement of our method over the best baseline.}\label{tb:main1}
\end{table*}

\begin{table*}[ht]
    \centering
    \begin{tabular}{ccccccccc}
    \hline
        Methods & Ref. & PIE & Scene & Leaves & NUS-WIDE & MSRC & Fashion \\ 
        \hline
        TMC & ICLR'21 & 17.79$\pm$12.19 & 18.19$\pm$3.82 & 21.00$\pm$5.20 & 15.42$\pm$4.05 & 72.38$\pm$17.92 & 31.28$\pm$0.86 \\ 
        ETMC & TPAMI'22 & 40.29$\pm$19.39 & 13.51$\pm$3.36 & 73.44$\pm$17.91 & 16.71$\pm$7.95 & 83.81$\pm$5.30 & 74.94$\pm$0.41 \\ 
        DUANets & AAAI'21 & 0.59$\pm$0.29 & 1.07$\pm$0.36 & 0.38$\pm$0.61 & 1.58$\pm$0.50 & 2.86$\pm$1.78 & 1.74$\pm$1.13 \\ 
        MMD & CVPR'22 & 11.18$\pm$14.96 & 4.48$\pm$2.85 & 0.60$\pm$0.73 & 0.21$\pm$0.19 & 38.10$\pm$46.66 & 4.22$\pm$2.51 \\ 
        QMF & ICML'23 & 18.82$\pm$7.52 & 7.47$\pm$3.10 & 22.75$\pm$2.26 & 10.83$\pm$3.22 & 70.00$\pm$23.11 & 15.60$\pm$0.46 \\ 
        ECML & AAAI'24 & 41.62$\pm$9.92 & 6.67$\pm$3.02 & 54.40$\pm$11.10 & 16.83$\pm$8.12 & 80.00$\pm$19.08 & 18.28$\pm$2.99 \\ 
        TMNR & IJCAI'24 & \underline{73.68$\pm$10.78} & 25.89$\pm$14.64 & 27.94$\pm$31.56 & 16.71$\pm$6.48 & 76.19$\pm$11.76 & 42.64$\pm$1.10 \\ 
        PDF & ICML'24 & 11.28$\pm$1.34 & 11.33$\pm$1.54 & 5.88$\pm$1.92 & 11.58$\pm$0.80 & 59.52$\pm$26.21 & 15.32$\pm$0.87 \\ 
        \midrule
        ETMC+AT & - & 1.47+0.81 & \underline{42.74+6.35} & \underline{79.44+1.89} & 27.04+2.86 & 80.95+9.76 & 20.36+7.35 \\
        ECML+AT & - & 7.50+11.07 & 33.00+4.87 & 78.81+7.68 & 25.38+4.36 & 60.48+29.30 & \underline{90.70+1.33} \\
        TMNR+AT & - & 71.32+11.44 & 38.08+16.64 & 49.62+3.39 & 20.33+2.28 & 73.81+3.01 & 72.16+0.69 \\
        PDF+AT & - & 49.85+9.12 & 32.37+1.59 & 54.38+2.49 & \underline{28.83+2.55} & \underline{81.43+6.63} & 77.26+0.79 \\
        \midrule
        RDML & Ours & \textbf{88.97$\pm$4.08} & \textbf{48.67$\pm$3.97} & \textbf{87.25$\pm$5.90} & \textbf{29.83$\pm$3.99} & \textbf{89.05$\pm$7.62} & \textbf{91.19$\pm$0.69} \\ 
        $\bigtriangleup$\% &  & \textcolor{blue}{+15.29} & \textcolor{blue}{+5.93} & \textcolor{blue}{+7.81} & \textcolor{blue}{+1.00} & \textcolor{blue}{+7.62} & \textcolor{blue}{+0.49} \\ 
    \hline
    \end{tabular}
    \caption{Classification accuracy (\%) under adversarial attacks, where only a random view is attacked. AT denotes adversarial training.}\label{tb:main2}
\end{table*}

\section{Experiments}
\subsection{Datasets and Competitors}
To verify the effectiveness and robustness of our method, we conduct experiments on six multi-view datasets, including PIE \cite{pie}, Scene \cite{scene}, Leaves \cite{leaves}, NUS-WIDE \cite{nus-wide}, MSRC \cite{msrc}, and Fashion \cite{fashion}. In addition, we compare our RDML method with eight state-of-the-art multi-view learning methods, including four evidence-based methods (i.e., TMC \cite{enn1}, ETMC \cite{enn7}, ECML \cite{enn2}, and TMNR \cite{enn3}), and four other uncertainty/confidence based methods (i.e., DUANets \cite{mv4}, MMD \cite{mmd}, QMF \cite{qmf}, and PDF \cite{pdf}. The details of all datasets and methods are shown in Appendix.

\begin{figure*}[htbp]
	\centering
	\begin{minipage}{0.195\linewidth}
		\centering
		\includegraphics[width=\linewidth]{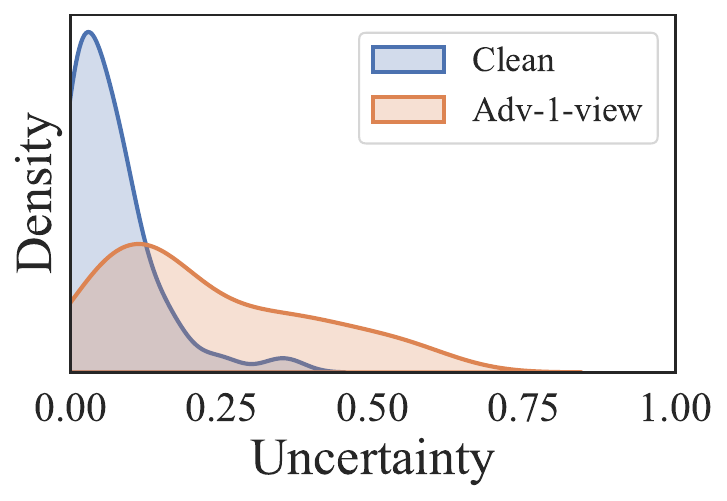}
	\end{minipage}
	\begin{minipage}{0.195\linewidth}
		\centering
		\includegraphics[width=\linewidth]{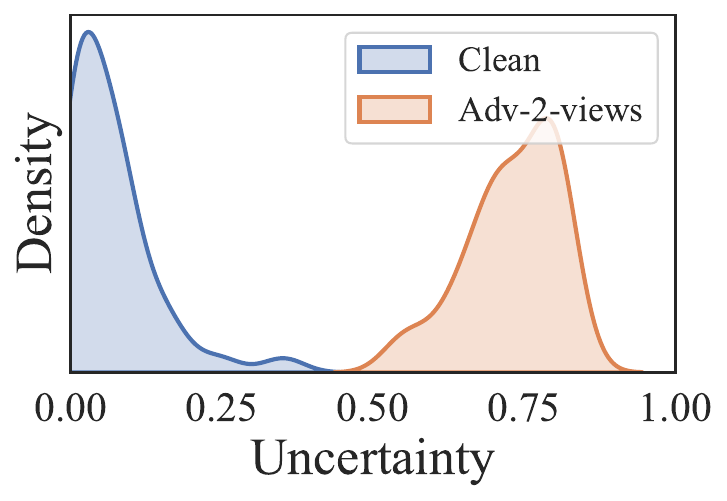}
	\end{minipage}
	\begin{minipage}{0.195\linewidth}
		\centering
		\includegraphics[width=\linewidth]{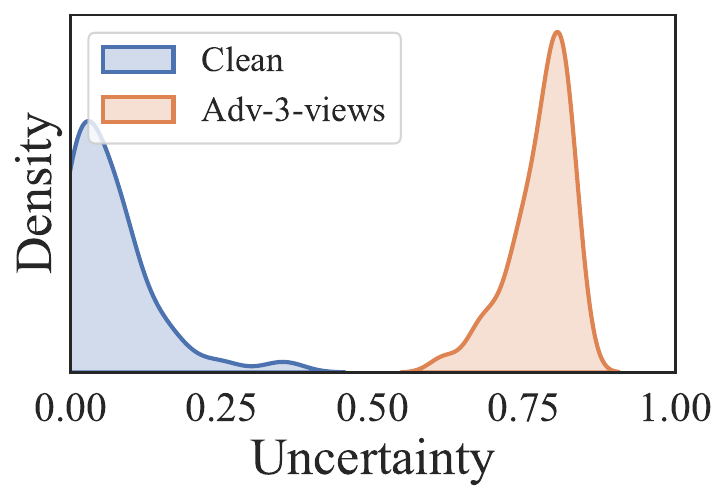}
	\end{minipage}
	\begin{minipage}{0.195\linewidth}
		\centering
		\includegraphics[width=\linewidth]{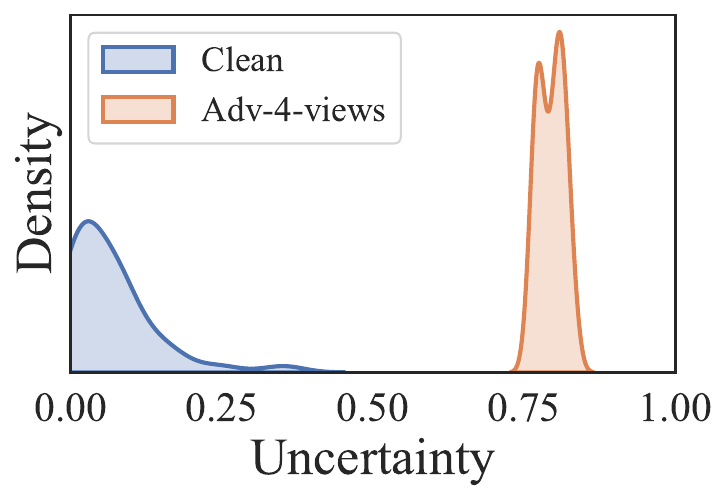}
	\end{minipage}
	\begin{minipage}{0.195\linewidth}
		\centering
		\includegraphics[width=\linewidth]{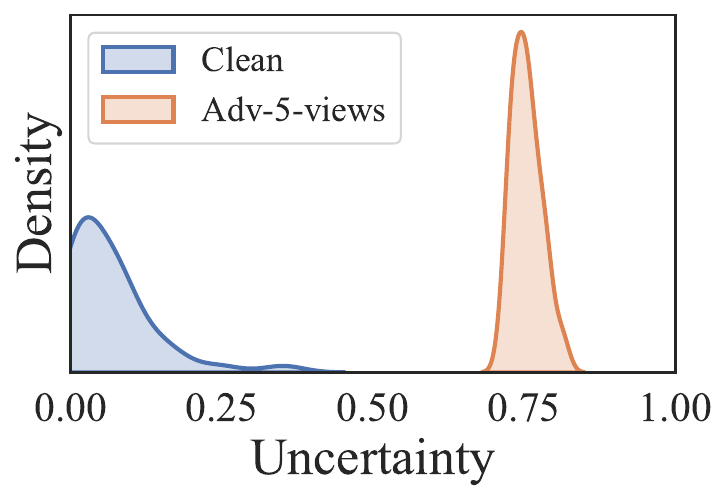}
	\end{minipage}
    \caption{Density of estimated uncertainty on MSRC with different numbers of attacked views.}
	\label{fig:u}
\end{figure*}
\subsection{Implementation Details}
Our experiments are conducted based on the PyTorch 2.4.1 framework with an Nvidia RTX 3090 GPU. For all datasets, 80\% of the samples are used for training (for our method, these data are also used for pretraining), and 20\% of the samples are used for testing. All experiments will be run 5 times, and we will report the average performance and standard deviation based on the accuracy of each test (after the last training epoch). The pretraining epoch of $E_{pt}(\cdot)$ is 1000 with a batch size of 500, and the training epoch is 500 for the cleaning setting and 400 for the adversarial setting. The learning rate is selected from $[0.003, 0.005]$. $E_{pt}^{v}(\cdot)$ and $E_{c}^{v}(\cdot)$ are with the size of $[d_{v},K]$. $f_{EM}^{v}$ is with the size of $[K,d_{v}]$. And $f_{EM}^{v}$ is with the size of $[d_{v},d_{v}]$. Adam is used as the optimizer. The temperature $\mu$ of Gumbel softmax is set as 0.1. We use Projected Gradient Descent for random view attack. The number of attack iterations is 10 with a maximum perturbation range of $8/255$.

\begin{figure}[htb]
\centering
\subfigure{\includegraphics[width=0.49\linewidth]{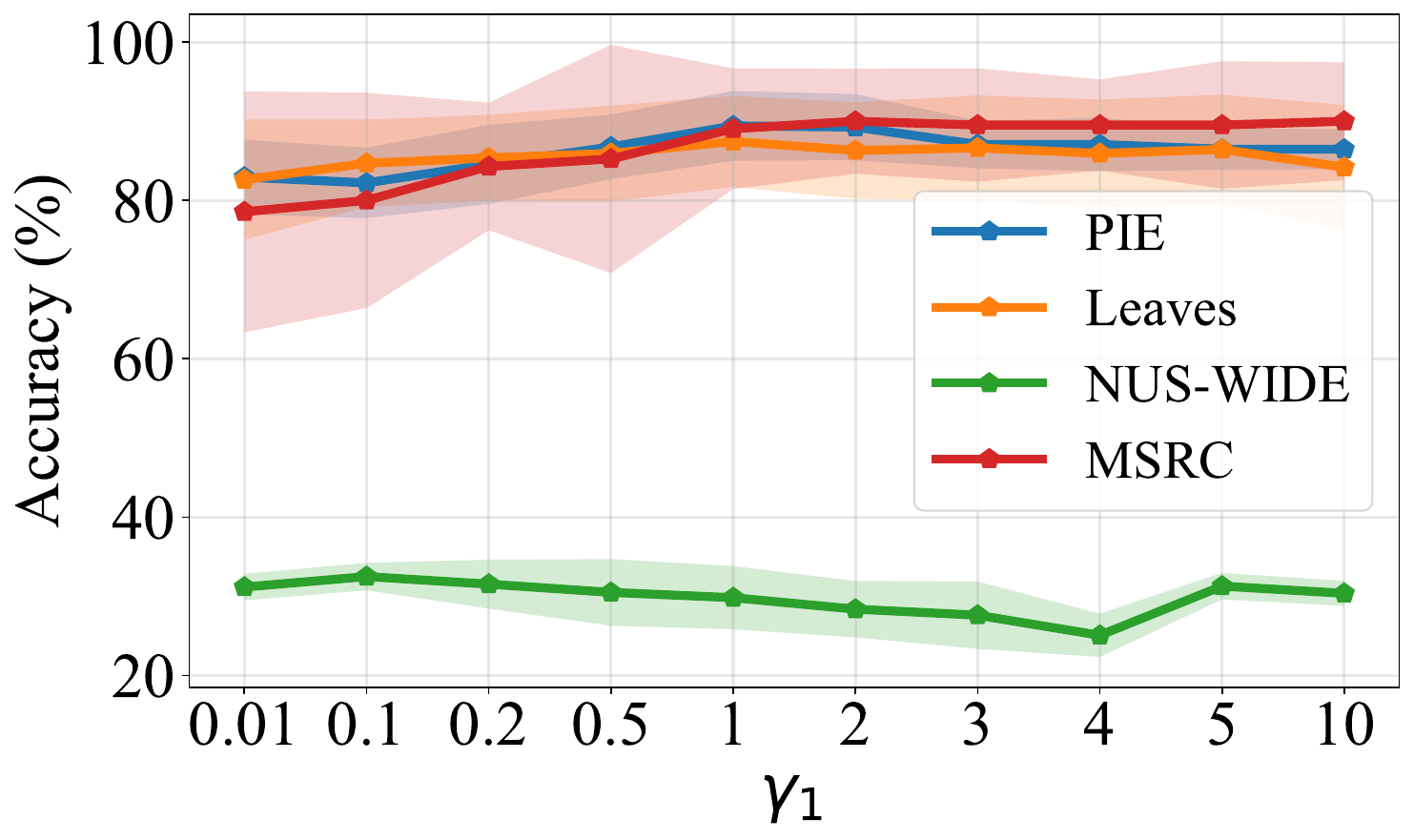}}
\subfigure{\includegraphics[width=0.49\linewidth]{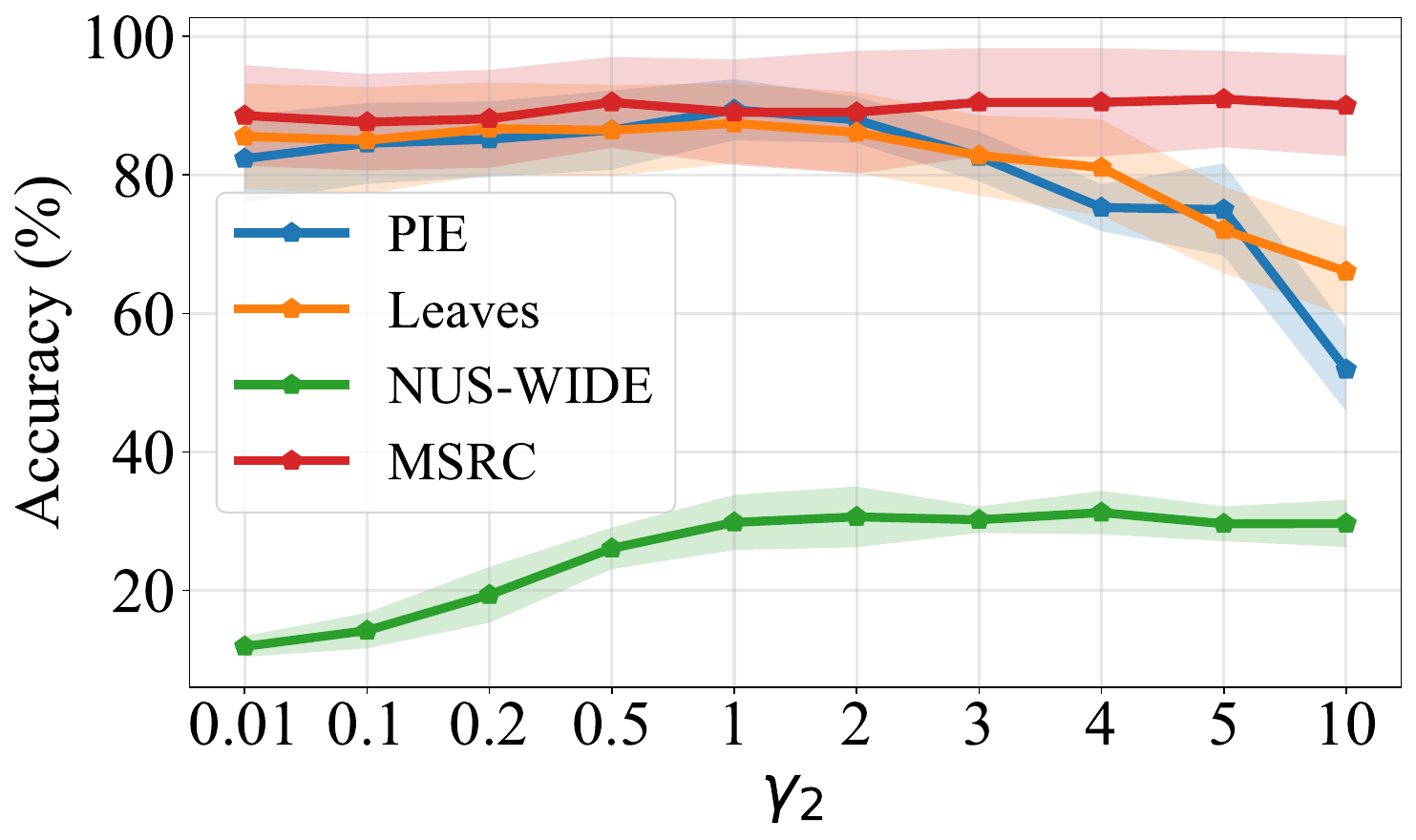}}
\caption{Classification accuracy (\%) on four datasets with different $\gamma_{1}$ and $\gamma_{2}$ (one random view is attacked).}
\label{fig:hpp}
\end{figure}

\subsection{Experimental Results}
To comprehensively evaluate the robustness of all methods, we conduct experiments on clean data and under adversarial attacks respectively. According to Table \ref{tb:main1} and Table \ref{tb:main2}, we have following observations.
\begin{itemize}
	\item{For clean multi-view data, RDML achieves the best performance in most cases, with an average improvement of 1.41\% compared to the best baselines. We attribute this to two reasons. First, the pretrained evidence extractor can provide good parameter initialization for the evidential classifier, avoiding falling into local optima prematurely during the training process. Second, evidential disentanglement learning, especially the combination of evidence-based disentanglement and View-specific evidential attention mechanism, is proficient in extracting multi-view features that are beneficial for classification and ignoring the interference of redundant information.}
	\item{RDML demonstrates significant advantages in AMVL. RDML has an average improvement of 6.36\% compared to the second best methods (including AT based methods). Unlike existing evidence learning methods, RDML obtains a robust evidence extractor through perturbation-insensitive pretraining. Moreover, evidence-based disentanglement is adept at separating clean and adversarial features. Weak adversarial features are then repaired by the feature recalibration module, while the View-specific evidential attention can shield the interference of hard adversarial features on multi-view classification. Therefore, RDML is effective and robust under adversarial perturbations.}
    \item{Both evidence based methods and other types of methods are highly vulnerable to adversarial attacks. Although adversarial training can, to some extent, relieve the sensitivity of multi-view models to adversarial attacks in many cases, due to the lack of an appropriate feature disentanglement mechanism, a large amount of adversarial and uninformative features are aggregated during multi-view fusion process, which impairs the classification performance.}
\end{itemize}

\subsection{Uncertainty Analysis}
To study the effectiveness of the uncertainty estimation mechanism of our method under adversarial perturbation, we visualize the estimated uncertainty density in more adversarial scenarios. As we can observe from Fig.\ref{fig:u}, when only a few views are attacked, the estimated uncertainty can be well matched with that in the clean setting. However, when more views are attacked, the estimated uncertainty gradually increases. This phenomenon, on the one hand, indicates that the uncertainty estimation mechanism of our method is effective and robust under adversarial conditions. On the other hand, it also reveals that the uncertainty estimation mechanism is affected by the quality of the views. The worse the view quality, the more difficult it is to make accurate decisions. This further demonstrates the effectiveness of our evidential disentanglement learning and feature recalibration module, which are designed to decouple and repair adversarial and informative features, significantly improving the view quality.

\subsection{Parameter Sensitivity Analysis}
Here we study the influence of two key hyperparameters, $\gamma_{1}$ and $\gamma_{2}$, on model robustness. There are two points we can get from Fig.\ref{fig:hpp}. (1) In most cases, the accuracy rises slowly as two values increase. It reaches the optimal performance when the values are 1 or 2, and then declines slowly. Overall, our method is insensitive to parameter changes, demonstrating its robustness. (2) When the values of $\gamma_{1}$ and $\gamma_{2}$ are too small (0.01) or too large (10), the accuracy decreases significantly. On the one hand, this further validates the effectiveness of the evidential disentanglement loss and feature recalibration loss ($\gamma_{1}$ and $\gamma_{2}$ are their respective balancing coefficients). On the other hand, it also shows that overemphasizing either decoupling adversarial features or repairing adversarial information will undermine the robustness. Striking an appropriate balance is the key to achieving better performance.

\begin{table}
\resizebox{1.0\linewidth}{!}{
\begin{tabular}{ccccccc|ccc}
    \hline
        \multicolumn{7}{c}{Ablation Module} & \multicolumn{3}{|c}{Dataset} \\ \hline
        $E_{pt}(\cdot)$ & ED & FC & ATT & $\mathcal{L}_{ACL}$ & $\mathcal{L}_{EDL}$ & $\mathcal{L}_{FRL}$ & PIE & Scene & Leaves \\ \hline
        - & \Checkmark & \Checkmark & - & \Checkmark & \Checkmark & \Checkmark & 51.62$\pm$14.46 & 48.34$\pm$5.14 & 67.56$\pm$5.03 \\ 
        \Checkmark & - & \Checkmark & \Checkmark & \Checkmark & - & \Checkmark & 5.44$\pm$3.50 & 46.13$\pm$5.85 & 11.31$\pm$6.30 \\ 
        \Checkmark & \Checkmark & - & \Checkmark & \Checkmark & \Checkmark & - & 82.79$\pm$2.35 & 41.58$\pm$6.91 & 84.25$\pm$6.92 \\ 
        \Checkmark & \Checkmark & \Checkmark & - & \Checkmark & \Checkmark & \Checkmark & 87.79$\pm$4.12 & 48.18$\pm$3.72 & 86.06$\pm$5.88 \\ 
        \Checkmark & \Checkmark & \Checkmark & \Checkmark & - & \Checkmark & \Checkmark & 86.47$\pm$3.80 & 47.54$\pm$4.55 & 86.81$\pm$4.92 \\ 
        \Checkmark & \Checkmark & \Checkmark & \Checkmark & \Checkmark & - & \Checkmark & 83.68$\pm$4.14 & 48.16$\pm$4.30 & 83.13$\pm$6.34 \\ 
        \Checkmark & \Checkmark & \Checkmark & \Checkmark & \Checkmark & \Checkmark & - & 81.76$\pm$6.21 & 46.98$\pm$4.70 & 86.06$\pm$7.58 \\ 
        \Checkmark & \Checkmark & \Checkmark & \Checkmark & \Checkmark & \Checkmark & \Checkmark & \textbf{88.97$\pm$4.08} & \textbf{48.67$\pm$3.97} & \textbf{87.25$\pm$5.90} \\ 
    \hline
\end{tabular}
}\caption{Ablation experiments (classification accuracy (\%)) on PIE, Scene, and Leaves with one randomly attacked view. ED, FC, and ATT denote evidential disentanglement learning, feature recalibration, and evidential attention mechanism, respectively.}
\label{tb:abl}
\end{table}

\subsection{Ablation Study}
We conduct ablation experiments to verify the effectiveness of key components of RDML. As shown in Table \ref{tb:abl}, after removing the pretrained $E_{pt}(\cdot)$ and evidence-based disentanglement respectively (the evidence attention mechanism and $\mathcal{L}_{EDL}$ then get invalid automatically), the model robustness significantly decreases. This indicates that: (1) perturbation-insensitive pretraining is highly effective in helping the evidential neural network resist adversarial attacks. (2) evidence-based decoupling can effectively strip the adversarial perturbations from view representations, and effectively reduce the interference of adversarial and uninformative features. After removing the feature recalibration and evidential attention respectively, the model performance shows a decline to varying degrees in most cases. In addition, we also explore the effectiveness of three proposed losses. The results prove that the three losses, especially $\mathcal{L}_{EDL}$ and $\mathcal{L}_{FRL}$, are able to improve the model robustness in most cases.

\section{Conclusion}
In this work, we reveal and study the AUP in trusted multi-view learning. To this end, we propose a novel Reliable Disentanglement Multi-view Learning framework. Specifically, RDML designs an evidential disentanglement learning to separate clean and adversarial features, and this process is guided by a pretrained evidence extractor. To mitigate the interference of adversarial features on multi-view decision, an adversarial feature recalibration module and an evidential attention mechanism are proposed. Experiments conducted on six datasets show the effectiveness and robustness of RDML against view adversarial attacks.

\section*{Acknowledgments}
This work is supported in part by the National Natural Science Foundation of China (Grant no. 62372315), Sichuan Science and Technology Planning Project (Grants nos. 2024NSFTD0049, 2024ZDZX0004, 2024YFHZ0089), the Chengdu Science and Technology Project (Grant no. 2023-XT00-00004-GX), and the Sichuan Science and Technology Miaozi Program (Grant no. MZGC20240057).
\appendix





\bibliographystyle{named}
\bibliography{ijcai25}

\begin{thebibliography}{}

\bibitem[\protect\citeauthoryear{Cao \bgroup \em et al.\egroup }{2024}]{pdf}
Bing Cao, Yinan Xia, Yi~Ding, Changqing Zhang, and Qinghua Hu.
\newblock Predictive dynamic fusion.
\newblock {\em International Conference on Machine Learning}, 2024.

\bibitem[\protect\citeauthoryear{Chua \bgroup \em et al.\egroup }{2009}]{nus-wide}
Tat-Seng Chua, Jinhui Tang, Richang Hong, Haojie Li, Zhiping Luo, and Yantao Zheng.
\newblock Nus-wide: a real-world web image database from national university of singapore.
\newblock In {\em Proceedings of the ACM international conference on image and video retrieval}, pages 1--9, 2009.

\bibitem[\protect\citeauthoryear{Cope \bgroup \em et al.\egroup }{2013}]{leaves}
James Cope, Thibaut Beghin, Paolo Remagnino, and Sarah Barman.
\newblock {One-hundred plant species leaves data set}.
\newblock UCI Machine Learning Repository, 2013.
\newblock {DOI}: https://doi.org/10.24432/C5RG76.

\bibitem[\protect\citeauthoryear{Fei-Fei and Perona}{2005}]{scene}
Li~Fei-Fei and Pietro Perona.
\newblock A bayesian hierarchical model for learning natural scene categories.
\newblock In {\em 2005 IEEE computer society conference on computer vision and pattern recognition (CVPR'05)}, volume~2, pages 524--531. IEEE, 2005.

\bibitem[\protect\citeauthoryear{Geng \bgroup \em et al.\egroup }{2021}]{mv4}
Yu~Geng, Zongbo Han, Changqing Zhang, and Qinghua Hu.
\newblock Uncertainty-aware multi-view representation learning.
\newblock In {\em Proceedings of the AAAI Conference on Artificial Intelligence}, volume~35, pages 7545--7553, 2021.

\bibitem[\protect\citeauthoryear{Goodfellow \bgroup \em et al.\egroup }{2015}]{atk4}
Ian~J. Goodfellow, Jonathon Shlens, and Christian Szegedy.
\newblock Explaining and harnessing adversarial examples.
\newblock In {\em International Conference on Learning Representations}, 2015.

\bibitem[\protect\citeauthoryear{Gross \bgroup \em et al.\egroup }{2010}]{pie}
Ralph Gross, Iain Matthews, Jeffrey Cohn, Takeo Kanade, and Simon Baker.
\newblock Multi-pie.
\newblock {\em Image and vision computing}, 28(5):807--813, 2010.

\bibitem[\protect\citeauthoryear{Han \bgroup \em et al.\egroup }{2020}]{enn1}
Zongbo Han, Changqing Zhang, Huazhu Fu, and Joey~Tianyi Zhou.
\newblock Trusted multi-view classification.
\newblock In {\em International Conference on Learning Representations}, 2020.

\bibitem[\protect\citeauthoryear{Han \bgroup \em et al.\egroup }{2022a}]{mmd}
Zongbo Han, Fan Yang, Junzhou Huang, Changqing Zhang, and Jianhua Yao.
\newblock Multimodal dynamics: Dynamical fusion for trustworthy multimodal classification.
\newblock In {\em Proceedings of the IEEE/CVF conference on computer vision and pattern recognition}, pages 20707--20717, 2022.

\bibitem[\protect\citeauthoryear{Han \bgroup \em et al.\egroup }{2022b}]{enn7}
Zongbo Han, Changqing Zhang, Huazhu Fu, and Joey~Tianyi Zhou.
\newblock Trusted multi-view classification with dynamic evidential fusion.
\newblock {\em IEEE transactions on pattern analysis and machine intelligence}, 45(2):2551--2566, 2022.

\bibitem[\protect\citeauthoryear{Ilyas \bgroup \em et al.\egroup }{2019}]{atk5}
Andrew Ilyas, Shibani Santurkar, Dimitris Tsipras, Logan Engstrom, Brandon Tran, and Aleksander Madry.
\newblock Adversarial examples are not bugs, they are features.
\newblock {\em Advances in neural information processing systems}, 32, 2019.

\bibitem[\protect\citeauthoryear{Jang \bgroup \em et al.\egroup }{2022}]{gbsm}
Eric Jang, Shixiang Gu, and Ben Poole.
\newblock Categorical reparameterization with gumbel-softmax.
\newblock In {\em International Conference on Learning Representations}, 2022.

\bibitem[\protect\citeauthoryear{Jsang}{2018}]{edl2}
Audun Jsang.
\newblock {\em Subjective Logic: A formalism for reasoning under uncertainty}.
\newblock Springer Publishing Company, Incorporated, 2018.

\bibitem[\protect\citeauthoryear{Kim \bgroup \em et al.\egroup }{2023}]{atk6}
Woo~Jae Kim, Yoonki Cho, Junsik Jung, and Sung-Eui Yoon.
\newblock Feature separation and recalibration for adversarial robustness.
\newblock In {\em Proceedings of the IEEE/CVF Conference on Computer Vision and Pattern Recognition}, pages 8183--8192, 2023.

\bibitem[\protect\citeauthoryear{Kurakin \bgroup \em et al.\egroup }{2018}]{atk7}
Alexey Kurakin, Ian~J Goodfellow, and Samy Bengio.
\newblock Adversarial examples in the physical world.
\newblock In {\em Artificial intelligence safety and security}, pages 99--112. Chapman and Hall/CRC, 2018.

\bibitem[\protect\citeauthoryear{Li \bgroup \em et al.\egroup }{2025}]{mv14}
Xingfeng Li, Yuangang Pan, Yuan Sun, Quansen Sun, Yinghui Sun, Ivor~W. Tsang, and Zhenwen Ren.
\newblock Incomplete multi-view clustering with paired and balanced dynamic anchor learning.
\newblock {\em IEEE Transactions on Multimedia}, 27:1486--1497, 2025.

\bibitem[\protect\citeauthoryear{Liang \bgroup \em et al.\egroup }{2024}]{kg}
Ke~Liang, Lingyuan Meng, Hao Li, Meng Liu, Siwei Wang, Sihang Zhou, Xinwang Liu, and Kunlun He.
\newblock Mgksite: Multi-modal knowledge-driven site selection via intra and inter-modal graph fusion.
\newblock {\em IEEE Transactions on Multimedia}, 2024.

\bibitem[\protect\citeauthoryear{Liu \bgroup \em et al.\egroup }{2022}]{enn6}
Wei Liu, Xiaodong Yue, Yufei Chen, and Thierry Denoeux.
\newblock Trusted multi-view deep learning with opinion aggregation.
\newblock In {\em Proceedings of the AAAI Conference on Artificial Intelligence}, volume~36, pages 7585--7593, 2022.

\bibitem[\protect\citeauthoryear{Liu \bgroup \em et al.\egroup }{2024}]{atk8}
Jun Liu, Jiantao Zhou, Jiandian Zeng, and Jinyu Tian.
\newblock Difattack: Query-efficient black-box adversarial attack via disentangled feature space.
\newblock In {\em Proceedings of the AAAI Conference on Artificial Intelligence}, volume~38, pages 3666--3674, 2024.

\bibitem[\protect\citeauthoryear{Long \bgroup \em et al.\egroup }{2022}]{atk2}
Teng Long, Qi~Gao, Lili Xu, and Zhangbing Zhou.
\newblock A survey on adversarial attacks in computer vision: Taxonomy, visualization and future directions.
\newblock {\em Computers \& Security}, 121:102847, 2022.

\bibitem[\protect\citeauthoryear{Madry \bgroup \em et al.\egroup }{2018}]{atk1}
Aleksander Madry, Aleksandar Makelov, Ludwig Schmidt, Dimitris Tsipras, and Adrian Vladu.
\newblock Towards deep learning models resistant to adversarial attacks.
\newblock In {\em International Conference on Learning Representations}, 2018.

\bibitem[\protect\citeauthoryear{Peng \bgroup \em et al.\egroup }{2019}]{mv6}
Xi~Peng, Zhenyu Huang, Jiancheng Lv, Hongyuan Zhu, and Joey~Tianyi Zhou.
\newblock Comic: Multi-view clustering without parameter selection.
\newblock In {\em International conference on machine learning}, pages 5092--5101. PMLR, 2019.

\bibitem[\protect\citeauthoryear{Qin \bgroup \em et al.\egroup }{2024}]{mv2}
Yalan Qin, Xinpeng Zhang, Shui Yu, and Guorui Feng.
\newblock A survey on representation learning for multi-view data.
\newblock {\em Neural Networks}, page 106842, 2024.

\bibitem[\protect\citeauthoryear{Sensoy \bgroup \em et al.\egroup }{2018}]{edl1}
Murat Sensoy, Lance Kaplan, and Melih Kandemir.
\newblock Evidential deep learning to quantify classification uncertainty.
\newblock {\em Advances in neural information processing systems}, 31, 2018.

\bibitem[\protect\citeauthoryear{Sun \bgroup \em et al.\egroup }{2024}]{mv7}
Yuan Sun, Yang Qin, Yongxiang Li, Dezhong Peng, Xi~Peng, and Peng Hu.
\newblock Robust multi-view clustering with noisy correspondence.
\newblock {\em IEEE Transactions on Knowledge and Data Engineering}, 2024.

\bibitem[\protect\citeauthoryear{Wang \bgroup \em et al.\egroup }{2022}]{atk3}
Jia Wang, Chengyu Wang, Qiuzhen Lin, Chengwen Luo, Chao Wu, and Jianqiang Li.
\newblock Adversarial attacks and defenses in deep learning for image recognition: A survey.
\newblock {\em Neurocomputing}, 514:162--181, 2022.

\bibitem[\protect\citeauthoryear{Wen \bgroup \em et al.\egroup }{2020}]{mv10}
Jie Wen, Yong Xu, and Hong Liu.
\newblock Incomplete multiview spectral clustering with adaptive graph learning.
\newblock {\em IEEE Transactions on Cybernetics}, 50(4):1418--1429, 2020.

\bibitem[\protect\citeauthoryear{Wen \bgroup \em et al.\egroup }{2021}]{mv9}
Jie Wen, Zheng Zhang, Zhao Zhang, Lunke Fei, and Meng Wang.
\newblock Generalized incomplete multiview clustering with flexible locality structure diffusion.
\newblock {\em IEEE Transactions on Cybernetics}, 51(1):101--114, 2021.

\bibitem[\protect\citeauthoryear{Wen \bgroup \em et al.\egroup }{2023}]{mv8}
Jie Wen, Zheng Zhang, Lunke Fei, Bob Zhang, Yong Xu, Zhao Zhang, and Jinxing Li.
\newblock A survey on incomplete multiview clustering.
\newblock {\em IEEE Transactions on Systems, Man, and Cybernetics: Systems}, 53(2):1136--1149, 2023.

\bibitem[\protect\citeauthoryear{Xiao \bgroup \em et al.\egroup }{2017}]{fashion}
Han Xiao, Kashif Rasul, and Roland Vollgraf.
\newblock Fashion-mnist: a novel image dataset for benchmarking machine learning algorithms.
\newblock {\em arXiv preprint arXiv:1708.07747}, 2017.

\bibitem[\protect\citeauthoryear{Xu \bgroup \em et al.\egroup }{2016}]{msrc}
Jinglin Xu, Junwei Han, and Feiping Nie.
\newblock Discriminatively embedded k-means for multi-view clustering.
\newblock In {\em Proceedings of the IEEE conference on computer vision and pattern recognition}, pages 5356--5364, 2016.

\bibitem[\protect\citeauthoryear{Xu \bgroup \em et al.\egroup }{2024a}]{enn2}
Cai Xu, Jiajun Si, Ziyu Guan, Wei Zhao, Yue Wu, and Xiyue Gao.
\newblock Reliable conflictive multi-view learning.
\newblock In {\em Proceedings of the AAAI Conference on Artificial Intelligence}, volume~38, pages 16129--16137, 2024.

\bibitem[\protect\citeauthoryear{Xu \bgroup \em et al.\egroup }{2024b}]{enn3}
Cai Xu, Yilin Zhang, Ziyu Guan, and Wei Zhao.
\newblock Trusted multi-view learning with label noise.
\newblock In {\em Proceedings of the 33rd International Joint Conference on Artificial Intelligence}, pages 5263--5271. ijcai.org, 2024.

\bibitem[\protect\citeauthoryear{Xu \bgroup \em et al.\egroup }{2024c}]{mv5}
Jie Xu, Yazhou Ren, Xiaolong Wang, Lei Feng, Zheng Zhang, Gang Niu, and Xiaofeng Zhu.
\newblock Investigating and mitigating the side effects of noisy views for self-supervised clustering algorithms in practical multi-view scenarios.
\newblock In {\em Proceedings of the IEEE/CVF Conference on Computer Vision and Pattern Recognition}, pages 22957--22966, 2024.

\bibitem[\protect\citeauthoryear{Xu \bgroup \em et al.\egroup }{2025}]{mv13}
Shilin Xu, Yuan Sun, Xingfeng Li, Siyuan Duan, Zhenwen Ren, Zheng Liu, and Dezhong Peng.
\newblock Noisy label calibration for multi-view classification.
\newblock In {\em Proceedings of the AAAI Conference on Artificial Intelligence}, pages 21797--21805, 2025.

\bibitem[\protect\citeauthoryear{Yuan \bgroup \em et al.\egroup }{2024}]{mv12}
Honglin Yuan, Shiyun Lai, Xingfeng Li, Jian Dai, Yuan Sun, and Zhenwen Ren.
\newblock Robust prototype completion for incomplete multi-view clustering.
\newblock In {\em Proceedings of the 32nd {ACM} International Conference on Multimedia}, pages 10402--10411. {ACM}, 2024.

\bibitem[\protect\citeauthoryear{Yuan \bgroup \em et al.\egroup }{2025}]{mv11}
Honglin Yuan, Yuan Sun, Fei Zhou, Jing Wen, Shihua Yuan, Xiaojian You, and Zhenwen Ren.
\newblock Prototype matching learning for incomplete multi-view clustering.
\newblock {\em IEEE Transactions on Image Processing}, 34:828--841, 2025.

\bibitem[\protect\citeauthoryear{Yue \bgroup \em et al.\egroup }{2025}]{enn10}
Xiaodong Yue, Zhicheng Dong, Yufei Chen, and Shaorong Xie.
\newblock Evidential dissonance measure in robust multi-view classification to resist adversarial attack.
\newblock {\em Information Fusion}, 113:102605, 2025.

\bibitem[\protect\citeauthoryear{Zhang \bgroup \em et al.\egroup }{2020}]{mv3}
Changqing Zhang, Yajie Cui, Zongbo Han, Joey~Tianyi Zhou, Huazhu Fu, and Qinghua Hu.
\newblock Deep partial multi-view learning.
\newblock {\em IEEE transactions on pattern analysis and machine intelligence}, 44(5):2402--2415, 2020.

\bibitem[\protect\citeauthoryear{Zhang \bgroup \em et al.\egroup }{2023}]{qmf}
Qingyang Zhang, Haitao Wu, Changqing Zhang, Qinghua Hu, Huazhu Fu, Joey~Tianyi Zhou, and Xi~Peng.
\newblock Provable dynamic fusion for low-quality multimodal data.
\newblock In {\em International conference on machine learning}, pages 41753--41769. PMLR, 2023.

\bibitem[\protect\citeauthoryear{Zhang \bgroup \em et al.\egroup }{2024}]{atk9}
Yufeng Zhang, Jianxing Yu, Yanghui Rao, Libin Zheng, Qinliang Su, Huaijie Zhu, and Jian Yin.
\newblock Domain adaptation for subjective induction questions answering on products by adversarial disentangled learning.
\newblock In {\em Proceedings of the 62nd Annual Meeting of the Association for Computational Linguistics}, pages 9074--9089, 2024.

\end{thebibliography}

\clearpage
\title{Supplementary Material of Reliable Disentanglement Multi-view Learning Against View Adversarial Attacks}
\section*{Appendix}
This supplementary material provides a comprehensive understanding of our RDML method. Specifically, we mainly introduce the algorithm procedure, the details of used datasets and compared methods, and more experimental analysis to support our research.

\section{Algorithm Procedure}
To better show the details of our proposed method, we give the workflow of our RDML in Algorithm \ref{alg.1}. 
\begin{algorithm}[!htb]
    \textbf{Input}: Multi-view data $x$; pretraining epoch $e_{pt}$ and training epoch $e_{t}$; evidence extractor $E_{pt}(\cdot)$ and $E_{c}(\cdot)$; evidence mapping layer $f_{EM}(\cdot)$ and evidential classifier $f_{EC}(\cdot)$\\
    \textbf{Output}: Joint opinion.
    \begin{algorithmic}[1] 
        \STATE \textcolor{blue}{...............................Pretraining stage...............................}
        \STATE Randomly initialize the parameters of $E_{pt}(\cdot)$.
        \WHILE{not reach the last epoch $e_{pt}$}
        \STATE Construct the adversarial instance with random view attacks by Eq. (1);
        \STATE Train the evidence extractor $E_{pt}(\cdot)$ with mixed data for evidence-based multi-view classification using Eq. (2).
        \ENDWHILE
        \STATE Share the parameters of $E_{pt}(\cdot)$ with $E_{c}(\cdot)$, and freeze the parameters of $E_{pt}(\cdot)$.
        \STATE \textcolor{magenta}{.................................Training stage................................}
        \STATE Randomly initialize the parameters of $f_{EM}(\cdot)$ and $f_{EC}(\cdot)$.
        \WHILE{not reach the last epoch $e_{t}$}
        \STATE Construct the adversarial instance with random view attacks by Eq. (1);
        \STATE Decouple view features with the guidance of $E_{pt}(\cdot)$ via Eq. (9)-Eq. (12);
        \STATE Recalibrate view features using Eq. (14);
        \STATE Construct view-level evidential attention to further mitigate the interference of adversarial perturbations via Eq. (16) and Eq. (17);
        \STATE Construct joint opinion using Dempster combination rule via Eq. (18);
        \ENDWHILE
    \end{algorithmic}
    \caption{The workflow of RDML}
    \label{alg.1}
\end{algorithm}

\section{Experimental details}
\subsection{Datasets}
\begin{table}[h]
\centering
\begin{tabular}{cccc}
\hline
Dataset&Class&Size&Dimensionality \\
\hline
PIE&68&680&484/256/279 \\
Scene&15&4485&20/59/40 \\
Leaves&100&1600&64/64/64 \\
NUS-WIDE&12&2400&64/144/73/128/225 \\
MSRC&7&210&24/576/512/256/254 \\
Fashion&10&10000&784/784/784 \\
\hline
\end{tabular}
\caption{Details of datasets}\label{tbdata}
\end{table}

We evaluate the performance of our propose method on the following multi-view datasets. 
\textbf{PIE} encompasses 680 facial images sourced from 68 subjects. Three kinds of views (intensity, LBP and Gabor) are selected in our experments. \textbf{Scene} dataset comprises 4485 images, which are categorized into 15 distinct indoor and outdoor scene classes. Three types of features are selected: GIST, PHOG, and LBP. \textbf{Leaves} dataset is constituted by 1600 leaf samples, which are collected from 100 diverse plant species. We extract three types of features as views: shape descriptors, fine-scale edges, and texture histograms. \textbf{NUS-WIDE} dataset comprises 269648 images from 81 concepts. We select 200 images from each of the top 12 classes, with a total of five types of views: CH, CM, CORR, EDH and WT. \textbf{MSRC-v5} from Microsoft Research in Cambridge contains 210 images and 7 classes with coarse pixel-wise labeled images. Five types of views are extracted: CM, HOG, GIST, LBP and CENT. \textbf{Fashion} comprises of grayscale images of 70,000 fashion products from 10 categories. We sample 1000 images from each class, with a total of three types of views. Details of these datasets are shown in Table \ref{tbdata}.

\begin{table*}[t]
    \centering
    \begin{tabular}{c|ccc|ccc}
    \hline
        \multirow{2}*{Method} & \multicolumn{3}{c|}{PIE} & \multicolumn{3}{c}{MSRC} \\ \cline{2-7}
        ~ & $eps=8/255$ & $eps=0.05$ & $eps=0.1$ & $eps=8/255$ & $eps=0.05$ & $eps=0.1$ \\ \hline
        RDML & 88.97$\pm$4.08 & 88.24$\pm$4.88 & 94.26$\pm$1.18 & 89.05$\pm$7.62 & 88.10$\pm$7.38 & 89.52$\pm$5.55 \\ \hline
        TMNR & 73.68$\pm$10.78 & 68.97$\pm$15.92 & 59.12$\pm$22.26 & 76.19$\pm$11.76 & 70.48$\pm$18.36 & 73.33$\pm$15.36 \\ \hline
        PDF & 11.28$\pm$1.34 & 8.53$\pm$4.38 & 8.53$\pm$4.78 & 59.52$\pm$26.21 & 51.90$\pm$32.38 & 51.43$\pm$35.04 \\
    \hline
    \end{tabular}
    \caption{$eps$ is the maximum perturbation range. Larger values indicate stronger attacks. $8/255$ is the empirically default setting. One random view is attacked.}\label{tb:intense}
\end{table*}

\subsection{Baselines}
To verify the effectiveness and robustness of our method, we compare RDML with eight state-of-the-art multi-view methods. \textbf{TMC} applies evidence theory to multi-view learning, dynamically fusing various views at the evidence level. \textbf{ETMC} is an improved version of TMC. It enhances the performance of TMC by adding a new view which is the concatenation of all original views. \textbf{DUA-Nets} capitalizes on reversal networks to amalgamate the intrinsic information sourced from diverse views, ultimately transforming them into a unified representation. \textbf{MMD} dynamically evaluates feature and modality informativeness with specific strategies and induces a transparent fusion algorithm. \textbf{QMF} utilizes uncertainty-aware weighting and a sampling-based regularization technology to enhance correlation, aiming to achieve reliable and robust multimodal fusion. \textbf{ECML} proposes a new multi-view opinion fusion method and a conflict measurement method to solve the problem of aggregating conflicting opinions. \textbf{TMNR} proposes a noise correlation matrix, through which the Dirichlet parameters are updated to mitigate the interference of label noise. \textbf{PDF} derives the predictable Collaborative Belief with Mono- and Holo-Confidence to reduce the generalization error upper bound and further proposes a relative calibration strategy.

\section{Multi-view Attack Analysis}
\begin{figure}[htb]
	\centering
	\includegraphics[width=0.45\textwidth]{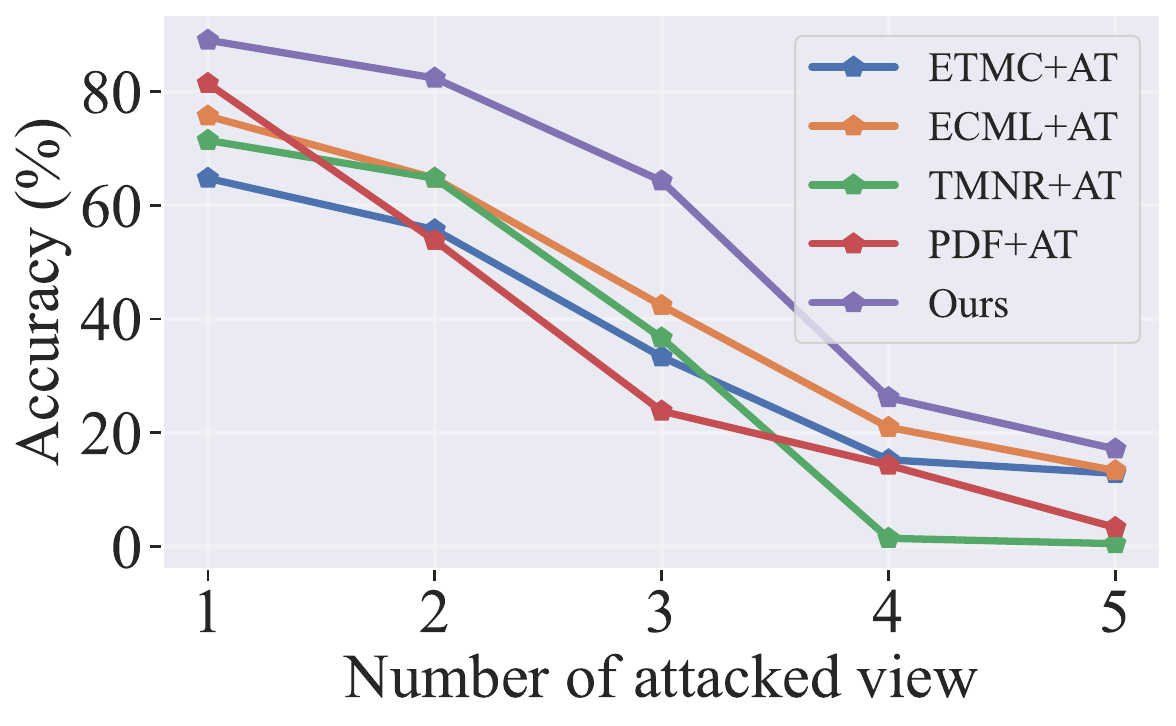}
	\caption{Classification accuracy (\%) on MSRC with different numbers of attacked views.}\label{fig:view_num}
\end{figure}
In order to further verify the robustness of our method, we conducted experiments in more difficult scenarios. As shown in Fig.\ref{fig:view_num}, as the number of attacked views increases, the performance of all methods is inevitably impaired, while our method can always maintain the best adversarial robustness. We attribute this to the combination of the adversarial-insensitive pretrained evidence extractor and the evidence-based disentanglement mechanism. The former can ensure the acquisition of effective evidences in a strong adversarial environment, and the latter can strip away adversarial perturbations and reduce their interference with the distribution of clean features.

\section{Attack Intensity} 
As shown in Table \ref{tb:intense}, TNMR and PDF exhibit clear performance degradation as the attack strength increases. Benefiting from perturbation-insensitive pretraining, RDML not only maintains robustness but even achieves superior performance under stronger attacks, further demonstrating its effectiveness.

\section{Parameter Freezing}
Since neural networks are highly sensitive to perturbations, we freeze the parameters of the evidence extractor after pretraining to ensure model stability and robustness. As shown in Table \ref{tb:freeze}, parameter freezing significantly enhances model robustness.
\begin{table}[h]
    \centering
    \begin{tabular}{c|ccc}
    \hline
        Method & PIE & Leaves & MSRC \\ \hline
        RDML & 88.97$\pm$4.08 & 87.25$\pm$5.90 & 89.05$\pm$7.62 \\ \hline
        w/o freezing & 78.24$\pm$7.03 & 84.56$\pm$6.33 & 86.67$\pm$8.73 \\
    \hline
    \end{tabular}
    \caption{Model performance with and without parameter freezing. One random view is attacked.}\label{tb:freeze}
\end{table}

\section{Gumbel Softmax}
In evidential disentanglement learning, we transform evidential representations into feature-level masks. To ensure differentiability, we introduce Gumbel softmax. So each value of the mask is non-negative and less than 1. The motivation of introducing the Gumbel distribution is to provide a differentiable reparameterization method for sampling discrete random variables, thereby addressing the issue that discrete variables cannot directly propagate gradients during backpropagation.

\end{document}